\documentclass[sigconf]{aamas} 

\usepackage{balance}

\usepackage[utf8]{inputenc} 
\usepackage[T1]{fontenc}    
\usepackage{hyperref}       
\usepackage{url}            
\usepackage{booktabs}       
\usepackage{amsfonts}       
\usepackage{nicefrac}       
\usepackage{microtype}      
\usepackage{xcolor}         
\usepackage{mathtools}
\usepackage{amsmath}
\usepackage{algorithmic}
\usepackage{graphicx}
\usepackage{algorithm}
\usepackage[utf8]{inputenc}
\usepackage{amsthm}
\usepackage[english]{babel}

\newtheorem{proposition}{Proposition}

\DeclareMathOperator*{\argmax}{arg\,max}

\setcopyright{none}
\acmConference[ALA '23]{Proc.\@ of the Adaptive and Learning Agents Workshop (ALA 2023)}
{May 29-30, 2023}{London, UK, \url{https://alaworkshop2023.github.io/}}{Cruz, Hayes, Wang, Yates (eds.)}
\copyrightyear{2023}
\acmYear{2023}
\acmDOI{}
\acmPrice{}
\acmISBN{}
\acmSubmissionID{2}

\title{Bandit-Based Policy Invariant Explicit Shaping for Incorporating External Advice in Reinforcement Learning} 
 
\author{Yash Satsangi}
\affiliation{
  \institution{JPMorgan Chase}
  \city{London}
  \country{United Kingdom}}
\email{yash.satsangi@jpmchase.com}

\author{Paniz Behboudian}
\affiliation{
  \institution{Huawei \& University of Alberta}
  \city{Edmonton}
  \country{Canada}}
\email{behboudi@ualberta.ca}

\begin{abstract}
A key challenge for a reinforcement learning (RL) agent is to incorporate external/expert\footnote{We use external advice and expert advice interchangeably in this paper.} advice in its learning. The desired goals of an algorithm that can shape the learning of an RL agent with external advice include (a) maintaining policy invariance; (b) accelerating the learning of the agent; and (c) learning from arbitrary advice \citep{behboudian20}. To address this challenge this paper formulates the problem of incorporating external advice in RL as a multi-armed bandit called shaping-bandits.  The reward of each arm of shaping bandits corresponds to the return obtained by following the expert or by following a default RL algorithm learning on the true environment reward.
We show that directly applying existing bandit and shaping algorithms that do not reason about the non-stationary nature of the underlying returns can lead to poor results. Thus we propose UCB-PIES (UPIES), Racing-PIES (RPIES), and Lazy PIES (LPIES) three different shaping algorithms built on different assumptions that reason about the long-term consequences of following the expert policy or the default RL algorithm. Our experiments in four different settings show that these proposed algorithms achieve the above-mentioned goals whereas the other algorithms fail to do so.
\end{abstract}

\keywords{Reward shaping, Reinforcement Learning, Non-stationarity}

\begin{document}

\pagestyle{fancy}
\fancyhead{}
\maketitle

\section{Introduction}

Reinforcement learning \citep{suttonbook} is a powerful paradigm for an agent to learn optimal behavior from sparse and delayed rewards. However, in many cases learning a good policy can consume significant time and resources \citep{vinyals2019alphastar}. For many real-world problems, external advice is available that can accelerate the learning of the agent. The practice of providing an RL agent with an additional reward to shape its learning is called \emph{reward shaping} and the external reward is called shaping reward \citep{gullapalli92,randlov98}. Incorporating external advice to shape the learning of an RL agent is a challenge because an external reward can distract the RL agent from its original goal and alter its optimal policy \citep{randlov98}. 

The challenge of incorporating external advice for RL agents depends on the \emph{nature} and \emph{form} of the external advice. In general, it is not possible to know the \emph{nature} of the external advice in advance, that is, whether the external advice is `good' (tells the agent optimal actions more often than not) or `bad' (leads more often to sub-optimal actions). The advice can be in any \emph{form}, that is, the advice could be available as a policy (a recommended action for each possible state) \cite{knox2012reinforcement} or as a value function (that gives value for each state), or as an arbitrary reward function. A fundamental problem with reward shaping is that incorporating arbitrary advice without any restrictions is prone to converging into suboptimal policies \citep{randlov98}. Potential-based reward shaping (PBRS) methods \citep{ng99,devlin12, wiewiora03} have been proposed as a well-established solution to this problem and are able to maintain policy invariance: they do not alter the optimal policy of the agent when incorporating external advice. However existing PBRS methods restrict the form of advice to potential functions that describe a `potential' for each state of the environment. By restricting the form of advice potential-based reward shaping methods \citep{ng99,devlin12} are able to maintain \emph{policy invariance}. While potential-based advice helps  maintain policy invariance, it can be infeasible for an expert or an agent to express advice as a potential function, especially for complex and large environments. It is much more natural for the expert to express the advice in an arbitrary reward form.  As formulated in \cite{behboudian20} a shaping algorithm ideally should satisfy the following requirements:
\begin{itemize}
    \item Policy invariance: keep the optimal policy unchanged \citep{ng99}
    \item Learn from arbitrary advice \citep{harutyunyan15}
    \item Accelerate learning of the agent if the advice is good 
\end{itemize}
Explicit shaping methods such as policy invariant explicit shaping (PIES) \citep{behboudian20} can learn from arbitrary advice and maintain policy invariance but its performance depends on a weighting parameter $\xi$ that determines the contribution of expert advice in the learning of the agent and decays according to a pre-determined schedule. Determining a good schedule of the decay parameters invariably requires the knowledge of the nature of the advice (good/bad) before the algorithm is run which is generally not possible. For example, if the advice is good, then decay the weight slowly and if it is bad then decay the weight quickly.

This paper builds on top of PIES and determines the contribution of advice online by formulting the problem of incorporating external advice in an RL agent as a multi-armed bandit problem (MAB) \citep{lattimore20}. More specifically, a 2-armed bandit called \emph{shaping bandit} where the agent must decide whether to follow the expert advice or follow a default RL policy that learns on true environment reward at every episode (or time step). We show that direct application of existing MAB algorithms such as $\epsilon$-greedy \citep{li19twolevel}, UCB \citep{auer02} or gradient-based optimization \citep{suttonbook,hu2020learning} to this problem can lead to poor results since these algorithms assume a stationary reward distribution for each arm. In the case of shaping bandits, the rewards are sampled depending on the expected return of the underlying RL agent that is still learning. Thus, the shaping algorithm must explicitly reason about the non-stationary nature of the returns obtained from the underlying RL agent. 

While exactly modeling the learning curve of an RL agent is a tedious task \citep{viering2021shape}, this paper starts with a simple assumption that the expected return from the underlying RL agent is monotonically increasing. Building on this assumption we propose a method to get upper and lower confidence bounds on the reward of arms for the shaping bandit. Next, based on different assumptions, we propose three different shaping algorithms lazy PIES (LPIES), upper confidence bound PIES (UPIES), and racing PIES (RPIES) that we show achieve all the goals of PIES and do not need to know the nature of the policy in advance. Additionally, we show that algorithms that do not reason about the non-stationary nature of the underlying RL agent's return can lead to policy invariance. We show these results in four different settings: a 2-armed bandit setting, a tabular RL navigation task, continuous control Cartpole task with a Deep RL agent, and on the game of Pong when a Deep-Q-Network (DQN) learns a policy directly from pixels. Our method is simple, easy to implement and RL algorithm agnostic that is it can be wrapped around any underlying RL agent irrespective of which RL algorithm the underlying RL agent is using.
Our results show that these shaping algorithms are able to exploit good advice to accelerate the learning of an RL agent, avoid bad external advice and still retrieve the optimal policy, and not get stuck on a sub-optimal policy even when advised by an expert. 

\section{Background}
An episodic MDP \citep{suttonbook} is given by the tuple $\langle S, A, T, R, H \rangle$ where $S $ is a finite set of states, $A$ is a finite set of actions, $T$ is the transition probabilities and $R$ is a reward function. At each time step, the environment is in a state $s \in S$, the agent takes an action $a \in A$, and the environment transitions to a new state $s' \in S$, according to the transition probabilities $T(s, a, s') = \Pr(s' | s, a)$. Additionally, the agent receives a reward whose expected value is given by the reward function $R(s,a)$. Finally, $H$ is a finite horizon.
A deterministic policy $\pi$ is a mapping from states to actions, $\pi : S \to A$, that is, for each, $s \in S$, $\pi(s)$ returns an action, $a = \pi(s)$. The \emph{$\tau$-step state-action value function} $Q_{\tau}^{\pi}(s,a)$ is defined as the expected sum of discounted rewards the agent will get if it takes action $a$ in state $s$ and follows the policy $\pi$ thereafter for $\tau$ steps until the horizon.
\begin{equation*}
Q_{\tau}^{\pi}(s,a) = \mathbb{E}[\sum_{k=0}^{\tau} R(s_{k},a_{k}) | s_0 = s, a_0 = a, \pi ]. 
\end{equation*}
The agent aims to find the optimal policy denoted by $\pi^*$ that maximizes the expected sum of rewards, and the state-action value function associated with $\pi^*$ is called the optimal state-action value function, denoted by $Q^{*}_{\tau}(s,a)$: $Q_{\tau}^*(s,a) = \max_{\pi \in \prod} Q_{\tau}^{\pi}(s,a), $ where $\prod$ is the set of all policies.
The action value function for a given policy $\pi$ satisfies the \emph{Bellman equation} \citep{putermanbook} 
\begin{equation*}
Q_{\tau}^{\pi}(s,a) = R(s,a) +  \mathbb{E}_{s',a'}[Q_{\tau-1}^{\pi}(s',a')] \ \forall \  \pi \in \prod,
\end{equation*}
where $s'$ and $a'$ are the state and action at the next time step.
The Bellman equation for the optimal policy $\pi^*$ is called the \emph{Bellman optimality equation}:
$Q_{\tau}^{*}(s,a) = R(s,a) + \mathbb{E}_{s',a'}[Q_{\tau-1}^{*}(s',a')].$
Given the optimal value function $Q_{\tau}^{*}(s,a)$, the agent can retrieve the optimal policy by acting greedily with respect to the optimal value function: $\pi^*(s) = \arg\max_{a \in A} Q_{\tau}^{*}(s,a).$


The idea behind many RL algorithms is to learn $Q^*_{\tau}$ iteratively. For example, SARSA~\citep{suttonbook} learns $Q$-values with the following update rule, at each time step $t$ ($Q_0$ can be initialized arbitrarily): 
\begin{equation}
Q_{\tau}(s,a) \gets  Q_{\tau}(s,a) + \alpha (r + Q_{\tau-1}(s',a')  - Q_{\tau}(s,a))  \nonumber
\label{eq:updaterule}
\end{equation}
where $\alpha$ denotes the learning rate, $Q_{\tau}$ denotes the estimates of $Q^*_{\tau}$ and $r$ is the sampled reward. In every episode, the agent follows an $\epsilon$-greedy policy with respect to $Q_{\tau}$.

\subsection{Policy Invariant Explicit Shaping (PIES)}
Policy invariant explicit shaping shapes the learning of an agent by explicitly learning two different value functions one on $R$, the true environment reward, and the other on $R^{expert}$, where $R^{expert}$ is a reward function specified by an expert.
\begin{equation}
\Phi_{\tau}(s,a) = \Phi_{\tau}(s,a) + \beta [R^{expert}(s,a) + \Phi_{\tau-1}(s',a') - \Phi_{\tau}(s,a) ] \nonumber
\end{equation}

The agent follows a policy $\pi^{*}(s) = \arg\max_{a \in A} Q^{*}_{\tau}(s,a) + \xi \Phi_{\tau}(s,a)$ where $\xi$ dies from 1 to 0 according to a pre-determined schedule. Since $\xi$ goes to zero after a certain fixed number of episodes, PIES is guaranteed to be policy invariant in limit \citep{singh00} and if the expert advice is good then the initial high weight on $\Phi$ leads to acceleration in the learning of the agent. However, since the schedule of $\xi$ is fixed, the agent may spend unnecessary resources (computations and samples) when the advice is bad, and in some cases not enough resources when the advice is good. 

\subsection{Upper Confidence Bound (UCB)}
Upper confidence bound \cite{auer02} algorithm describes a policy for decision-making in the stochastic multi-armed bandit (MAB) \cite{lai1985} problem. A stochastic $k-$armed bandit problem is defined by $k$ random variables $X_{m,t}$ for $1 \leq m \leq k$ and $t \geq 1$. At each time step $t$ the agent must select $m$ (or pull the arm $m$) and receives rewards $X_{m, t}$  that are independent and identically distributed according to an unknown law with an unknown expected value. The expected values of the rewards are assumed to be stationary with time.  The aim of the agent is to pull the arm with the maximum expected reward (over all time-steps).
The UCB policy is to pull the arm that maximizes $
\bar{X}_m + \sqrt{\frac{2 \log(M)}{M_j}}$, where $\bar{X}$ is the average reward received from arm $m$ for pulling it $M_j$ times, and $M$ is the total number of time steps.

\section{Problem Formulation \& Method}
\subsection{Shaping Bandits}
We consider an episodic MDP setting, where an agent learns by interacting with the environment in episodes. We formulate the problem of incorporating external advice as a 2-armed bandit that we call \emph{PIES-bandit} or \emph{shaping bandit}: at the start of each episode, the agent must select either to follow a \emph{default} policy that is $\epsilon$-greedy towards the values function $Q$ learned on true environment reward or it may choose to follow the expert advice that is the policy that is $\epsilon$-greedy towards $\Phi$. We call these two arms the $Q$ and $\Phi$ arms. At the end of the episode, the agent observes the cumulative return as the reward for pulling the $Q$ or the $\Phi$ arm. As such the aim of the agent is to pull the sequence of arms that gives the highest cumulative return over a period of $T$ episodes. 
Let $\mu: \{1, 2, \dots, T\} \to \{0,1\}$ denote a policy that returns a binary value in response to the episode number $i \in \{1, 2, \dots, T\}$. Here the binary value $\mu(i)$ represents the arm recommended by the policy and by default 0 corresponds to $Q$ arm and 1 corresponds to $\Phi$ arm. 
Let $X_{i,\mu} \in [0,1]$ denote the random variable that is the return obtained at the end of episode $i \in \{1, 2, \dots, T\}$ when following policy $\mu$. Ideally, the agent would like to follow $\mu^* = \arg\max_{\mu \in \{1,2,\dots,T\} \times \{0,1\}}J(\mu,T) $ where $J(\mu,T)$ is the expected sum of returns obtained for following $\mu$
\begin{equation}
J(\mu,T) = \mathbb{E}[\sum_{i=1}^{T} X_{i,\mu}]. \nonumber
\end{equation}
This is equivalent to minimizing the regret \citep{arora2012,heidari16} given by:
\begin{equation}
R(\mu,T) = J(\mu^*,T) - J(\mu,T).  \nonumber
\end{equation}
that is the difference between the expected sum of returns if the optimal policy $\mu^*$ was followed and a given policy $\mu$. For explicit shaping, by minimizing the regret the agent would be able to achieve policy invariance since the agent will ideally select the expert arm only if it leads to high returns over a period of time and accelerate learning of the agent (most likely, when the advice is good). If the advice is not good the agent will want to select the default arm since it yields a higher sum of expected returns.
In general, it is not possible to say if the regret can be minimized for the shaping bandits without making an assumption about the nature of $X_{i, \mu}$ \citep{arora2012}, which in turn depends on the underlying RL agents.

A common assumption regarding $X_{i,\mu}$ is that its expected value is stationary in which case this regret can be written as simple regret \citep{auer02}, and algorithms such as UCB can be used to minimize it. This assumption is not applicable to PIES-bandits since $X_{i, \mu}$ is the return of an RL agent that is learning, and its expected value may change with the number and sequence of the arms that are pulled. 
The above observation is important, especially for formulating explicit shaping as a bandit problem. Consider the following example, where the expert advice guides the agent (unknowingly) to a sub-optimal policy, let's say, that returns a percentage of the value of the optimal policy. If there was no expert advice then the RL agent is likely to find the optimal policy in limit \citep{singh00}.  However, since we would like to shape the learning of the agent according to the expert's advice it is imperative that the agent's learning will be slightly biased by the expert's policy. In that case, it is highly likely that in the initial episodes due to abundant feedback the expert policy would give high returns. A usual (say $\epsilon$-greedy or UCB or gradient-based) bandit policy in such case would keep pulling the $\Phi$ arm since it will give a high return from the start, whereas the $Q$ (default RL) arm will take multiple pulls before its average reward crosses the average reward provided by the $\Phi$ (expert) arm. As a result, the agent will follow the sub-optimal policy for a good number of episodes before (and if) it is able to find the optimal policy. In order to avoid this, the shaping algorithm must reason about the long-term consequences of the change in expected values of $X_{i,\mu}$ when deciding which arm to pull.
In the next subsections, we describe three different algorithms that are able to address the challenge of incorporating external advice in the learning of an RL agent and achieve the above mentioned goals of shaping. We first describe Lazy PIES which is able to achieve these goals without any restrictive assumptions. Next, we describe RPIES and UPIES that are able to achieve these goals under a couple of `strong' assumptions.

\subsection{Lazy PIES (LPIES)}
A starting point is to observe that the aim of an RL agent is to maximize its return, that is, to improve the policy until it yields the optimal value. Furthermore, since an RL agent learns from its mistakes, it is reasonable to assume that the more experience the agent gathers, the more improvement it can make to its policy, meaning that on average as the number of episodes increases, the expected return obtained from pulling either of the arms is going to go up (and most likely not down). Moreover, once the agent learns the optimal policy, it is unlikely, the agent will return to a sub-optimal policy. This intuition motivates us to propose a simple algorithm for shaping called  Lazy Policy Invariant Explicit Shaping (LPIES). LPIES follows a simple rule to pull between the $\Phi$ and $Q$ arm: pull with equal probability each arm until the average of historical returns from $\Phi$ arm is less than the average of historical returns from arm $Q$, then eliminate the $\Phi$ arm and keep pulling $Q$ (forever). Since the $Q$ arm is never eliminated the agent is guaranteed to converge to the optimal policy. The $\Phi$ arm will be eliminated as long as it is not optimal, in which case the agent will learn the optimal policy anyways but faster since the agent will follow the expert advice with probability 0.5 (in the case of two arms) and will be biased by the experience collected by it.

\subsection{Shaping as Monotone Bandit}
LPIES does not make any strong assumption about the learning of the underlying RL agent. However, when many experts are present then the acceleration offered by LPIES might not be the `fastest'. Furthermore, in some rare cases, LPIES may eliminate a good expert by chance (due to a bad sample). Next, we describe UPIES (and RPIES) that use confidence intervals on the value of pulling each arm in order to accelerate the learning of the agent more aggressively. In order to do so, we formulate the shaping bandit as a special case of the monotone bandit as described in \cite{heidari16}. To start with we assume that the expected rewards associated with the arms of PIES-bandits are monotone in the number of pulls of that arm. Second, the expected reward of an arm is independent of the number of pulls of other arms. To state this assumptions formally, let $n_O \in \{1,2, \dots, T \}$ denote the number of pulls of arm $O \in \{Q, \Phi\}$ and let $\rho_O(n_O)$ denote the expected rewards associated with arm $O$ after $n_O$ pulls. Then we assume $\rho_O(n_O) \geq \rho_O(n_O-1)$ for $O \in \{Q, \Phi \}$. Given this assumption, we can write $J(\mu, T) = \sum_{i=1}^{n_Q}\rho_Q(i) + \sum_{i=1}^{n_{\Phi}} \rho_{\Phi}(i)$. To tie $X_{i,\mu}$ to $\rho$, $X_{i,\mu}$ is drawn from an unknown distribution with the expected value $\rho_{O}(n_O)$, where $O = \mu(i)$ and $n_O = \sum_{j=0}^{i} \mathbb{I}({\mu(j), \mu(i)})$, where $\mathbb{I}({\mu(j), \mu(i)}) = 1$ if $\mu(i) = \mu(j)$ else 0.

Most importantly, these assumptions allow us to leverage proposition 1 from \citep{heidari16} that states that in an offline setting if arms of a bandit offer increasing rewards then there exists an arm $i^*$ such that the optimal policy is to pull $i^*$ for all the rounds. For PIES-bandits, this means that the optimal policy with the highest expected value is the one that pulls either the default RL ($Q$) arm or the expert arm ($\Phi$) for all episodes. 
\begin{proposition} (Haideri et al, 2016)
Assuming $\rho_O(n_O) \geq \rho_{O}(n_O - 1)$ for $O \in \{Q, \Phi\}$, then, the optimal policy \\ $\mu^{*} = \arg\max_{\mu \in \{1,2, \dots, T \} \times \{0, 1 \} } J(\mu, T)$ is to pull $\Phi$ (1) arm or $Q$ (0) arm for all $T$ episodes.
\end{proposition}

\citep{heidari16} prove the above proposition by contradiction. The main idea behind the proof is that for any policy $\mu$ that is not $\mu_Q$ or $\mu_{\Phi}$, it is possible to obtain a higher value than $\mu$ by replacing all pull of one arm with another (depending on whether $\rho(n_Q)$ is greater or lesser than $\rho(n_{\Phi})$). 

Thus, at any point, we are looking for a policy that in the future will only pull either the $Q$ arm or the $\Phi$ arm. Now, if only we can eliminate one of these arms, we will be able to decide which arm to pull for the rest of the episodes. 

\subsection{Racing PIES (RPIES)}
We leverage this result to propose a racing algorithm for PIES-bandits called Racing-PIES (RPIES). The main idea behind the RPIES is racing algorithm \citep{maron97} which is to pull all arms in a candidate set of arms in a round-robin fashion (or with equal probability) and eliminate arms from the candidate set if the upper confidence bounds of the arm are lower than the maximum of lower confidence bound across all arms. RPIES applies this idea to PIES-bandits in the following way. 
Let $\mu_{Q,t}$ and $\mu_{\Phi,t}$ denote policies such that for the rest of the episodes (after $t$) $\mu_{Q,t}$ recommends pulling only $Q$ arm and $\mu_{\Phi,t}$ recommends pulling only $\Phi$ arm. At any time step $t$, either $\mu_{Q,t}$ or $\mu_{\Phi,t}$ is the optimal policy for the rest of the rounds. RPIES tries to find that by maintaining upper and lower confidence bounds on $J(\mu_{Q,t}, T-t)$ (in short $J_Q$) and $J(\mu_{\Phi,t}, T-t)$ (in short, $J_\Phi$). RPIES uses separate deep neural networks to estimate the values of $J_{Q}$ and $J_{\Phi}$. These DNNs, specifically feed-forward neural networks $f_{w}$ are trained on the previous rewards obtained by pulling the $Q$ or the $\Phi$ arm. Furthermore, to enforce the monotone assumption the weights of the DNNs, $w$ are constrained to be non-negative. The DNNs are trained on a dataset that consists of the pull number and reward obtained for that pull for each of the $Q$ and $\Phi$ arm, for example, $D = \{ (1, r_1), (2, r_2) \dots (n, r_n) \}$. The DNN is trained by minimizing mean squared error. Once trained, the DNN is used to get an estimate of $\hat{J} = \frac{1}{T-t}\sum_{j=n}^{n+T-t} f_w(j)$. These estimates are divided by the total number of rounds remaining ($\frac{1}{T-t}$) to limit their values between 0 and 1 which makes it easier to apply Hoeffding's inequality \citep{hoeffding63} later.
For both arms, every time the arm is pulled, RPIES computes new estimates of $\hat{J}_{Q}$ and $J_{\Phi}$ by training a fresh neural network. Thus, if $n_Q$ denotes the number of pulls of arm $Q$, RPIES maintains a set of estimates of ${J}_{Q}$ of size $n_Q$, $\{\hat{J}_{Q,1}, \hat{J}_{Q,2} \dots, \hat{J}_{Q, n_Q}\}$ (same for $\Phi$ arm). RPIES treats these samples as independent and identically distributed (i.i.d) and obtains an upper confidence interval by applying Hoeffding's inequality \citep{hoeffding63} \footnote{Hoeffding's inequality: If $X_1,X_2, \dots X_n$ are independent and $0 \geq X_i \geq 1$ for all $i$ then for $t>0$,
\begin{equation}
\Pr(|X - \mathbb{E}X| \geq t)  \leq 2e^{-2nt^2},
\end{equation}
where $X = \frac{1}{n}(X_1 + X_2 + \dots + X_n)$.
}
Using Hoeffding's inequality on the sample, we get with probability
$1-\delta$
\begin{equation} \label{eq:ucb}
\frac{1}{n_Q}\sum_{i=1}^{n_Q}\hat{J}_{Q,i} + \sqrt{\frac{1}{2n_Q}\log(\frac{2}{\delta})} \geq  \frac{1}{T-t} \mathbb{E}J_Q
\end{equation}
At each time step, RPIES pulls the $Q$ or the $\Phi$ arm with probability 0.5 each until one of the arms is not eliminated. An arm is eliminated when its upper confidence bound is lower than the lower confidence bound of the other arm. 

Estimating a lower confidence bound does not require us to make use of the DNN. Instead, the monotone assumption makes it quite straightforward to draw a lower confidence interval on $J_Q$ and $J_\Phi$. Since the reward can only increase with the number of pulls the expected value of $J_Q$ per round can only be greater than the expected value of the average of previous rewards. For example for the $Q$ arm
\begin{equation}
\frac{1}{n_Q} \mathbb{E}  \sum_{i=1}^{n_Q} [X_{i, \mu_Q}] \leq \frac{1}{T-t}\mathbb{E}J_Q \nonumber
\end{equation}

The left-hand side of the above equation again can be lower bounded with probability $1-\delta$ using Hoeffding's inequality \citep{hoeffding63} 
\begin{equation}
\frac{1}{n_Q}\mathbb{E} \sum_{i=1}^{n_Q} [X_{i, \mu_Q}] \geq \frac{1}{n_Q} \sum_{i=1}^{n_Q} [X_{i, \mu_Q}]  - \sqrt{\frac{1}{2n_Q}\log(\frac{2}{\delta})} \nonumber
\end{equation}

It is now possible to leverage existing proof techniques \citep{maron97} to show that RPIES will eliminate the sub-optimal arm and hence be policy invariant. We bypass all of that by simply rigging the race in RPIES in favor of the $Q$ arm. This is because we know that the vanilla $Q$ arm is optimal in limit and we would not want to eliminate it. RPIES never eliminates the $Q$ arm instead, it waits until the upper confidence interval of $\Phi$ arm is lower than the lower confidence interval of $Q$ arm and then it eliminates $\Phi$. Note that the only way $\Phi$ is not eliminated is if $\Phi$ is indeed the optimal value function which is not bad. Thus, in the limit, RPIES is policy invariant by design.

In general, under this framework, any expert/external advice that is not the optimal policy will not survive the elimination. Instead, even nearly-optimal external policy will only distract the learning of the underlying RL agent, since the agent must follow a sub-optimal policy for a fixed set of episodes. However, in case the external advice is optimal, then the RPIES agent will be able to bias the experience of the underlying RL agent with the experience gained from following the optimal policy and this can help the RL agent to learn faster. Next, we propose UPIES having proposed a method for obtaining confidence intervals on the value of $Q$ and $\Phi$ arms.

\subsection{UCB-PIES (UPIES)}
We can now propose an upper confidence bound algorithm for PIES-Bandit, that at each round pulls the arm with the highest upper confidence bound given by: 
\begin{equation}
I = \argmax_{O \in \{Q, \Phi\}}    \frac{1}{n_{O}} \sum_{i=1}^{n_O} \hat{J}_{O} + \sqrt{\frac{2 \log(n_Q + n_\Phi)}{n_O}} \nonumber
\end{equation}

In the rest of the paper, we show empirically that the above two algorithms satisfy the goals of PIES but without the need to know the nature of the external advice in advance. We conduct experiments on a bandit problem first, followed by a tabular RL grid word setting and then a deep Q network applied to the cart pole environment.

\section{Experiments}
In this section, we first show that policies ($\epsilon$-greedy/UCB) that do not consider the long-term potential of the underlying RL agent can lead to poor results. We show this in a two-armed bandit setting and then in a grid world. Following this, we show the performance of LPIES, UPIES and RPIES when the advice is good and bad in a grid world. The final two experiments compare the performance of LPIES, UPIES, and RPIES to an RL agent that is learning without shaping on Cartpole and Pong. In these experiments, the underlying RL agent is using DNN-based function approximation, and the advice is coming from an external agent that has been trained to act in the respective environment.
\subsection{Two-armed bandit}
In the first experiment, we consider a simple two-armed bandit problem where arm 1 of the bandit offers a constant expected reward of 0.5  whereas arm 0 offers an expected reward that slowly but linearly increases from 0 to $Y \in \{0.05, 0.25, 0.75, 0.95\}$ (at the rate of 0.01 per pull). The agent's observations are corrupted by zero-mean Gaussian noise with a variance of 0.1. The purpose of this bandit is to simulate a setting where one arm is slowly learning towards a worse/better long-term reward (default RL agent) whereas the other arm is able to provide a constant reward from the start (expert arm). 
We compare the performance of UPIES against $\epsilon$-greedy and a version of UPIES where we train the neural network without the non-negative weight constraint. This baseline is called non-monotone-UPIES since it does not put constraints on the weights $w$ of the DNN. In general, we expect $\epsilon$-greedy to perform worse when $Y$ is greater than 0.5 and perform quite well when $Y$ is less than 0.5. In comparison, the performance of UPIES is expected to be good across all the settings. 

\begin{figure}
\begin{center}       
\includegraphics[scale=0.52]{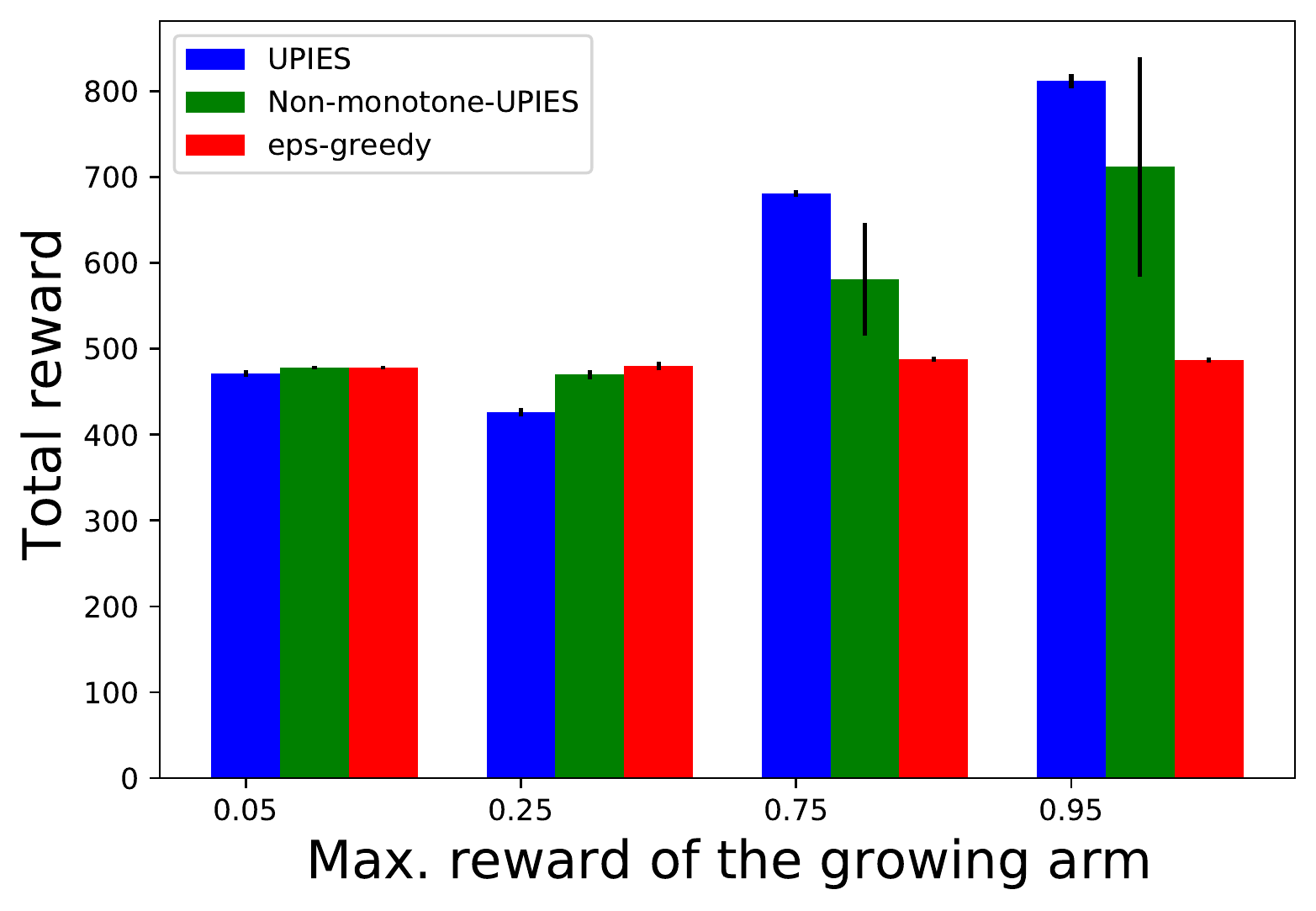}
\includegraphics[scale=0.6]{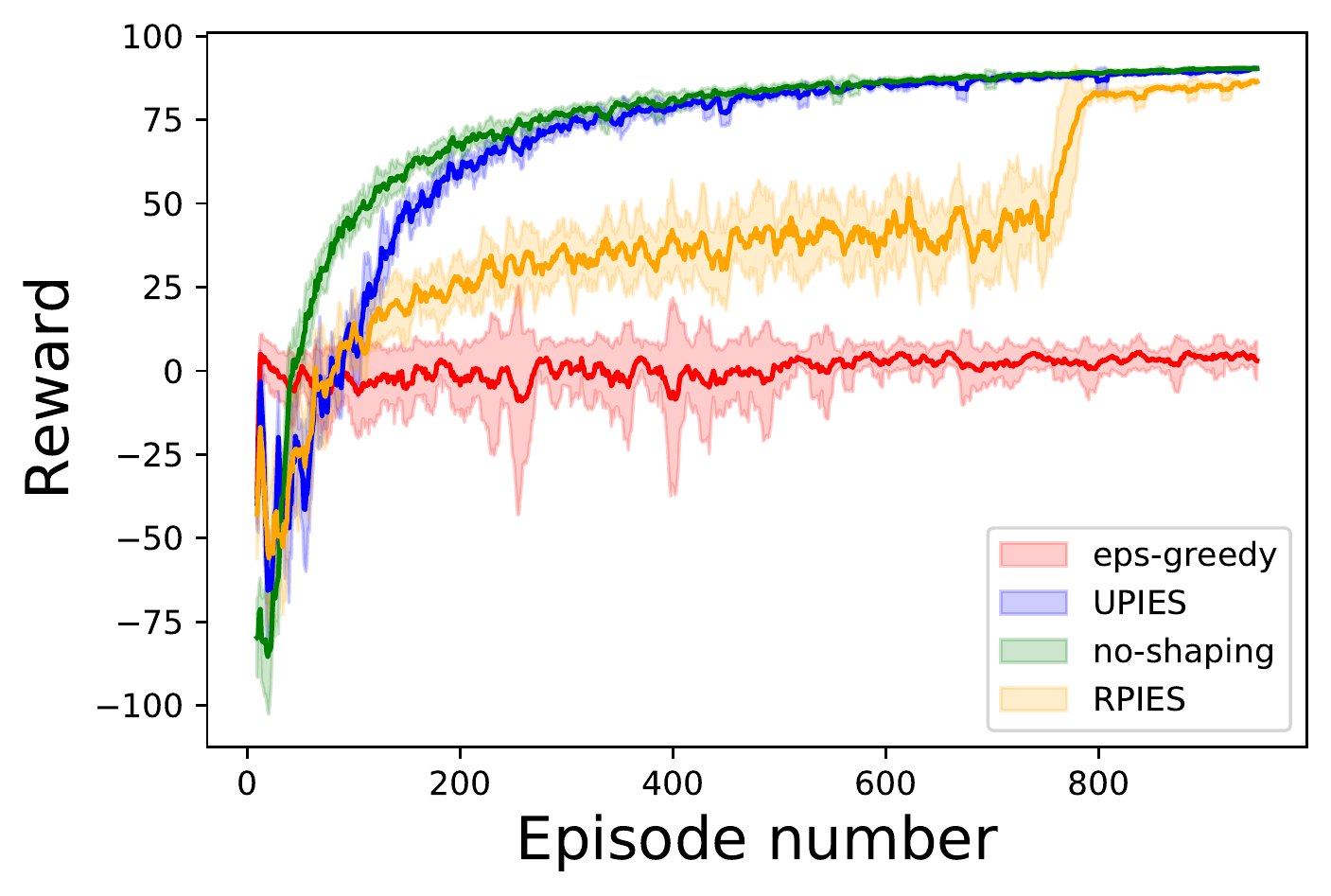}
\caption{(top) Results on a two-armed bandit problem. One of the arms yields a constant reward of 0.5 and the other arm yields a linearly increasing reward at the rate of 0.01 per pull with maximum reward in the set $\{0.05, 0.25, 0.75, 0.95\}$ 
(bottom) Results on a grid-world environment where the advice is not adversarial but sub-optimal and takes the agent to an easier but smaller reward. Bandit-based methods are able to retrieve the optimal policy whereas $\epsilon$-greedy policy prefers to pull the shaping arm since it offers relatively high rewards during the initial rewards. Y-axis is the return at the end of the episode. X-axis is the episode number.} \label{fig:bandit-experiment}
\end{center}
\end{figure}

Figure \ref{fig:bandit-experiment} shows the cumulative reward earned by the three policies over 1000 episodes across ten runs. As the value of $Y$ increases, we see that both $\epsilon$-greedy and non-monotone-UPIES fail to perform as well as UPIES.  Both these policies prefer the constant reward arm since it offers larger rewards in the initial pulls. While $\epsilon$-greedy invariably prefers the constant reward arm, the performance of the non-monotone-UPIES depends on the historical reward sample it receives. Since non-monotone-UPIES is free to fit a downward-looking curve to the rewards of arm 2 if a recent bad sample (consecutive low rewards) is observed, it leads to a significant decrease in the average cumulative reward for non-monotone UPIES. In cases when such a bad sample is not observed the algorithm identifies the upward-moving trend of the increasing arm and collects a good reward. This nature of this algorithm is highlighted by its high variance. 
UPIES is able to pick the best arm in all four cases. When the value of $Y$ is less than 0.5, it is able to minimize the performance gap between itself and greedy strategies such as epsilon greedy. However, as the value of $Y$ increases, the long-term reasoning of UPIES leads to larger cumulative rewards as compared to the other baselines. The lowest cumulative reward that UPIES achieves is for $Y = 0.25$, mainly due to pulling arm 0 until it is confident that it will not yield a larger reward than arm 1 in the future.  Finally, the performance difference between $\epsilon$-greedy and UPIES highlights the role of the count-based exploration bonus that UPIES uses, whereas the performance difference between non-monotone UPIES and UPIES highlights the role of monotone constraints on the weight of DNNs used by UPIES. 
 For all experiments, for all methods using bandits-based shaping (UPIES, RPIES and LPIES) we used a feed-forward DNN with three layers of size 8,4 and 1 respectively. The DNNs were trained for 2 epochs every time.

\subsection{Grid-world}
Next, we show experiments on an RL agent learning to navigate to a goal in a 20 x 20 grid-world \citep{suttonbook, behboudian20}. The agent starts at the coordinates (0,0) and the goal of the agent is located at (20,20). If the agent navigates to the goal, it receives a reward of 100 and the episode ends. At every time step, the agent must take an action and it pays a small negative reward of 0.1 (to encourage the agent to find the goal as quickly as possible). The episode ends after 2000 time steps if the agent is not able to find the goal state. We consider three different types of advice that can shape the learning of the agent: (a) good advice, (b) friendly but sub-optimal advice, and (c) adversarial advice. Good advice rewards the agent +0.1 every time it takes an `east' or `south' action, that is, it moves towards the goal. If the agent is to follow this advice then it will reach the goal state via one of the shortest paths and will earn the maximum reward. 
Adversarial advice offers the agent a reward of +0.1 for taking actions `west' or `north'. As such it recommends the agent do the exact opposite of the good advice. Needless to say, this is the worst advice that the agent can follow and it leads to large negative rewards. A `friendly but sub-optimal advice' rewards the agent +0.1 for taking action `east' but leads the agent to a sub-optimal state located at (20,0) which rewards the agent +5 and the episode ends. The point of friendly advice is that it represents advice that leads an agent to an easier but sub-optimal goal. The underlying RL agent follows the SARSA algorithm with optimistic initialization.

The expectation from a shaping algorithm is to perform well in all these three scenarios. While $\epsilon$-greedy policies will perform well when the advice is good or adversarial (on one of the extremes), it suffers when the advice leads to an easier but sub-optimal goal. Figure \ref{fig:bandit-experiment} (bottom) shows the performance of UPIES, RPIES, $\epsilon$-greedy algorithm and a no-shaping agent when the advice is friendly but sub-optimal. Similar to the 2-armed bandit setting, $\epsilon$-greedy policy is not able to let go sub-optimal friendly advice arm. This is mainly because the sub-optimal advice leads the underlying RL agent relatively quickly to a goal. The $\epsilon$-greedy policy is happy to collect the rewards from the friendly advice arm, not reasoning about the learning of the vanilla RL arm. 
Both UPIES and RPIES, on the other hand, are able to reason about the learning of the default RL arm. While UPIES is able to retrieve the optimal policy almost as quickly as the agent with no-shaping. RPIES on the other hand is cautious before eliminating the sub-optimal arm, nonetheless, it is able to retrieve the optimal policy. 

\begin{figure}
\begin{center}
\includegraphics[scale=0.6]{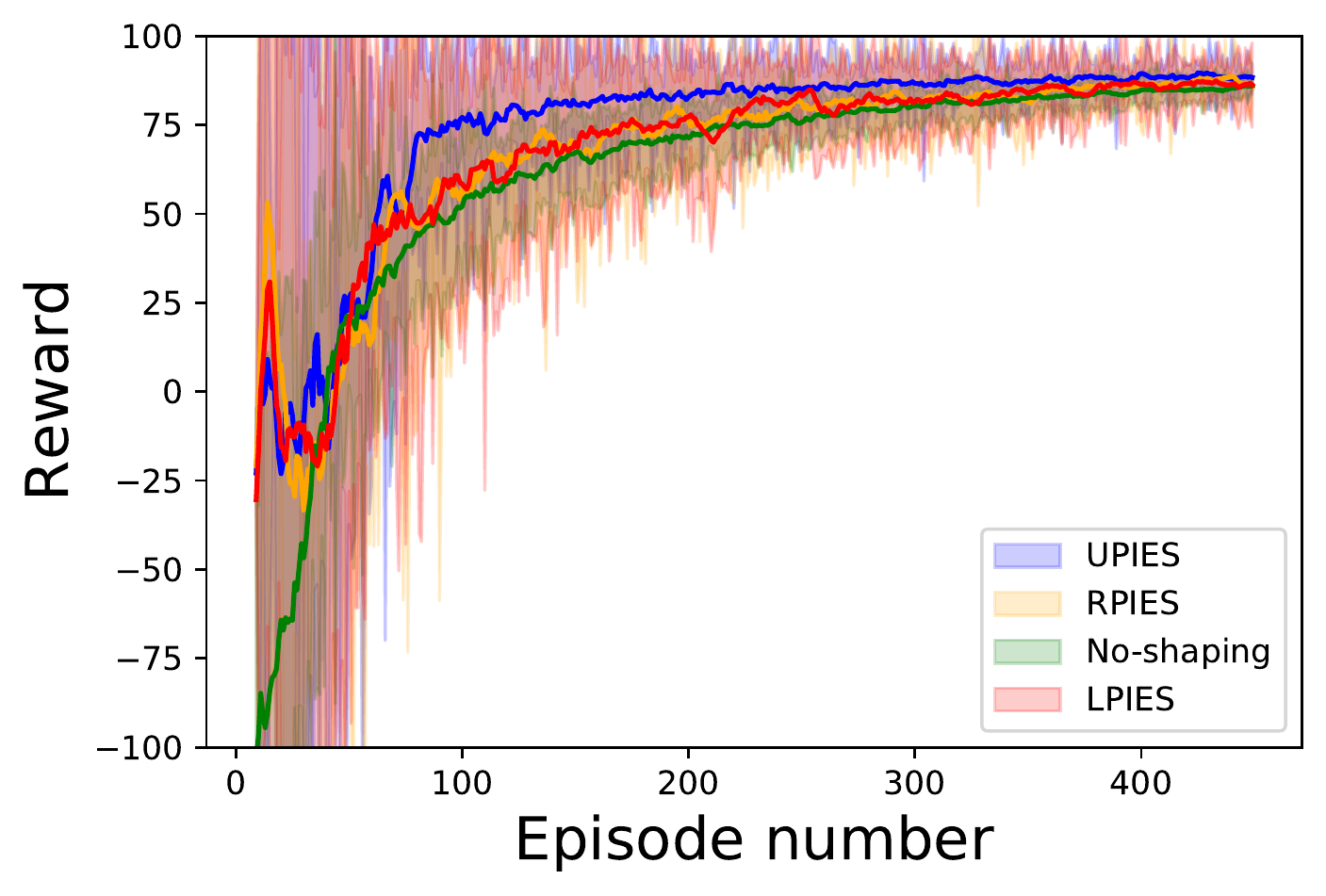}
\includegraphics[scale=0.6]{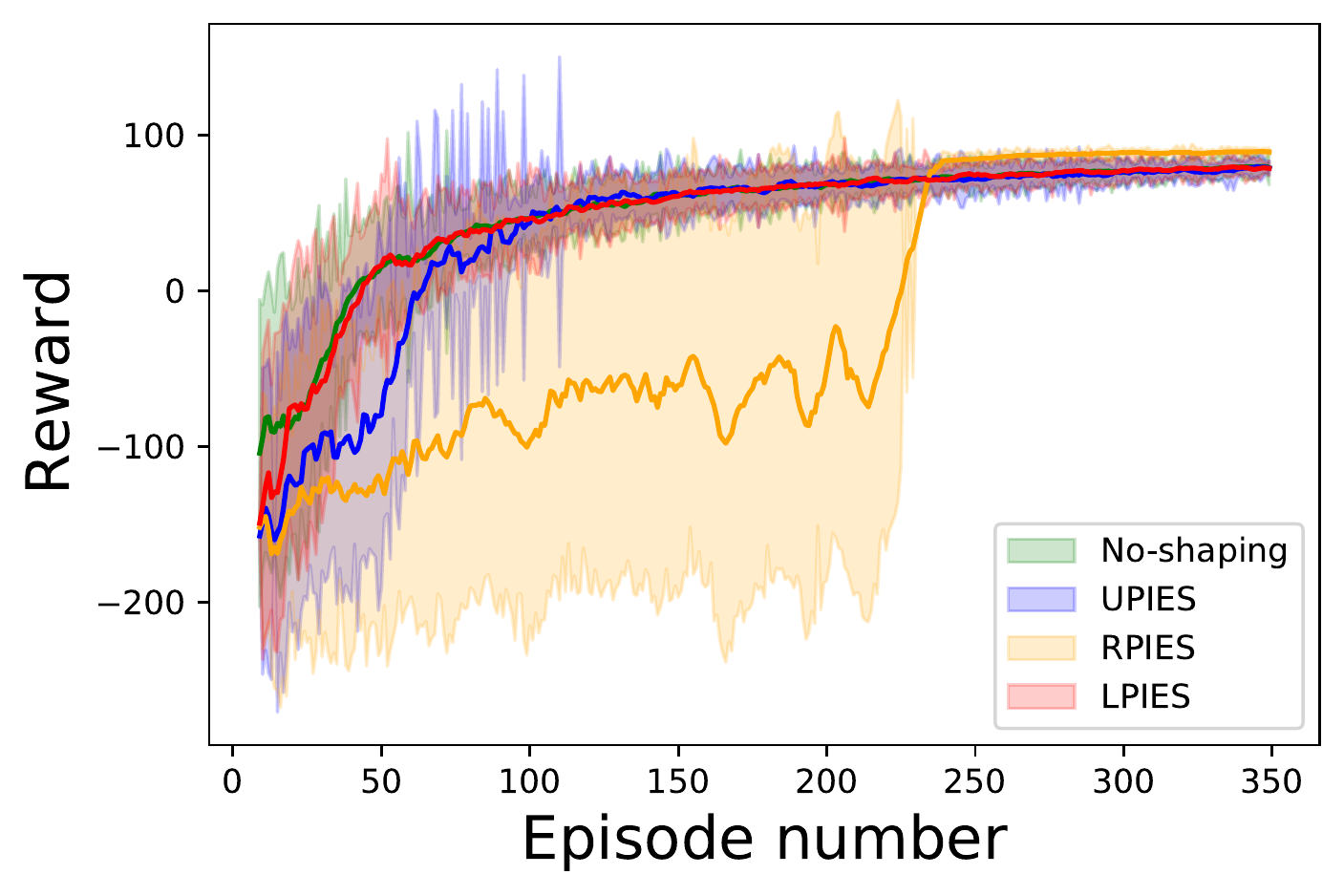}
\caption{ (top) Results on a grid-world environment where the advice is nearly optimal. Bandit-based agents are able to accelerate the learning of the agent. Y-axis is the return at the end of the episode. X-axis is the episode number.
(bottom) Results on a grid-world environment where the advice is adversarial (drives the agent away from the goal). Bandit-based agents are able to retrieve the optimal policy. Y-axis is the return at the end of the episode. X-axis is the episode number.} \label{fig:good}
\end{center}
\end{figure}

Next, we show that LPIES, UPIES and RPIES (collectively Bandit-PIES) accelerate the learning of the agent when the advice is good and they are quickly able to ignore the adversarial advice.
Figure \ref{fig:good} (top) shows the performance of Bandit-PIES when good advice is available. Bandit-PIES are able to accelerate the learning of the RL agent as compared to a no-shaping agent when the advice is good. Figure \ref{fig:good} (bottom) shows the performance of Bandit-PIES when the advice is adversarial. Both UPIES and RPIES get distracted slightly, however, again both algorithms are able to retrieve the optimal policy quite quickly. The high variance of RPIES is mainly due to the back-and-forth between the default RL arm and the adversarial RL arm. LPIES however is able to eliminate adversarial advice from the start.

The previous experiments demonstrate the need for long-term reasoning for the learning of the underlying RL agent when shaping the learning of the agent. Next, we show that Bandit-PIES are able to accelerate the learning of the agents when the advice is good (even if not optimal) when the underlying RL agent is using deep neural network-based function approximation.

\subsection{Cart-pole with deep neural networks}
\begin{figure}
\begin{center}
\includegraphics[scale=0.6]{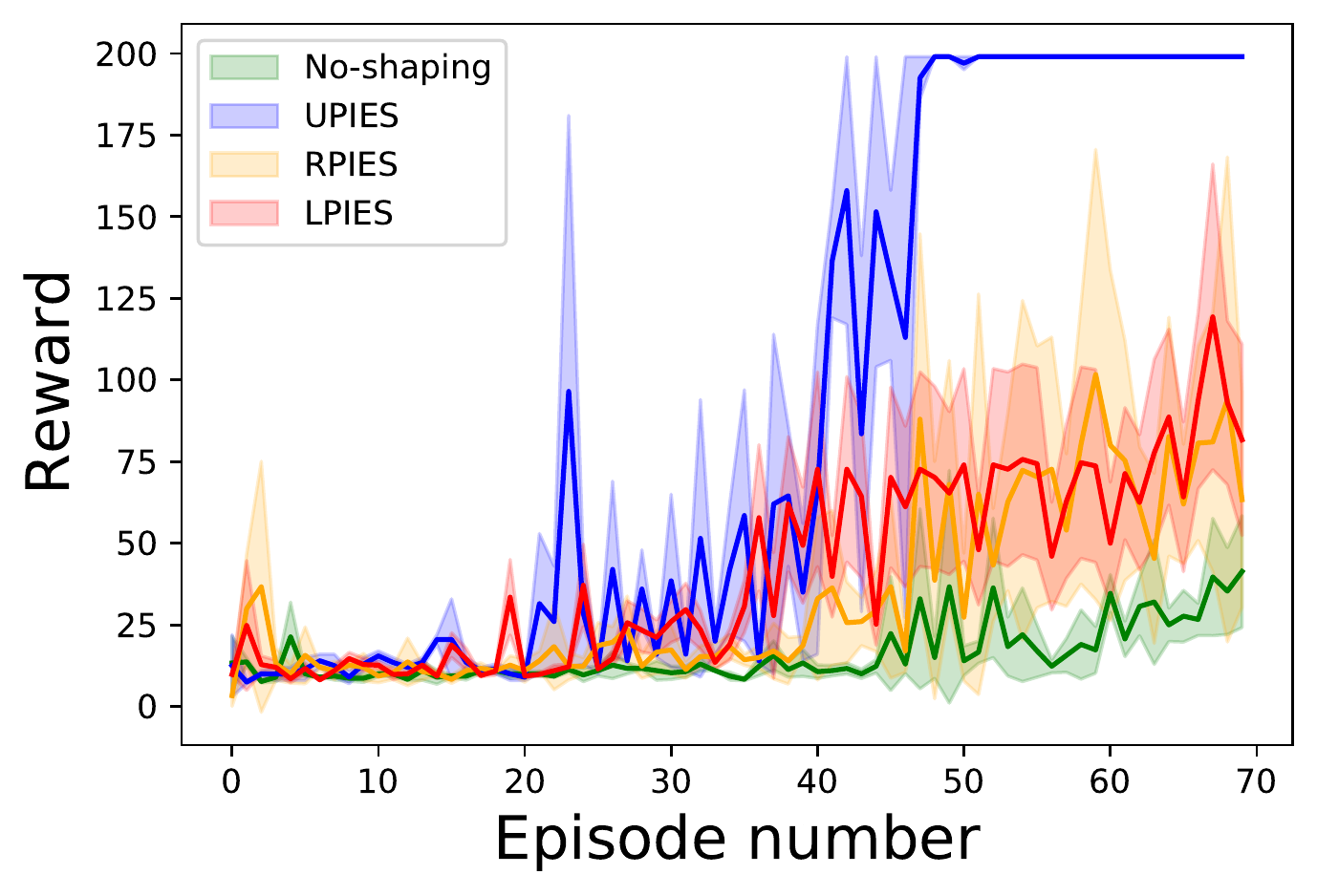}
\includegraphics[scale=0.6]{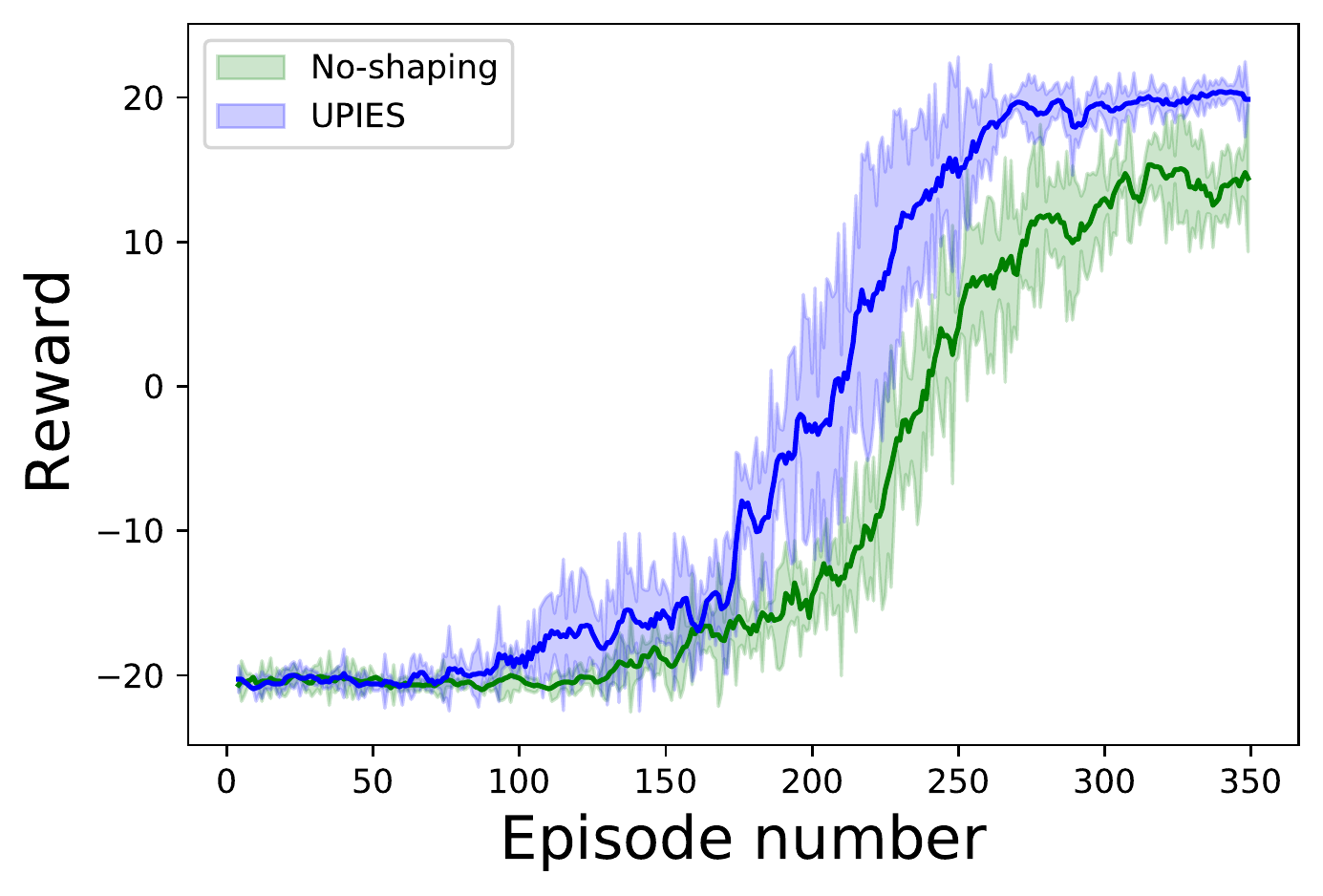}
\caption{{(top) Results on a cart pole environment where the advice is nearly-optimal. Bandit-based agents are able to converge to the optimal policy faster. Y-axis is the return at the end of the episode. X-axis is the episode number. 
(bottom) Results on pong where the advice comes from imitating a DQN trained to play Pong. Bandit-based agents are able to converge to the optimal policy faster. Y-axis is the return at the end of the episode. X-axis is the episode number.}} \label{fig:cartpole}
\end{center}
\end{figure}

In the next experiment, we show UPIES and RPIES are able to incorporate external advice in a deep RL agent. Furthermore, the advice, in this case, comes from another deep RL agent that is trained to near completion. The expert agent is a DQN that is trained to balance a cart-pole \citep{sutton88} to near completion. 
The untrained RL agent is also a DQN that tries to learn to balance cart pole from scratch. As the external advice, the agent gets a positive reward (+0.1) for taking the same action as the expert agent. Both the expert agent and the RL agent are DQNs with three layers of size 16, 16, and 2 respectively. The RL agent is trained with a learning rate of 0.001 using Adam \citep{kingma2014adam} optimizer. The value of gamma is set to 0.99 and the exploration $\epsilon$ starts at 0.1 and gradually decays to 0.01 over a period of 100 episodes.
In this case, we compare the performance of Bandit-PIES against an agent with no shaping. Since the expert agent is already trained to balance the cart pole, we expect the external advice to be good and accelerate the learning of the agent. Figure \ref{fig:cartpole} show the performance of Bandit-PIES against a no-shaping agent. While Bandit-PIES are able to accelerate the learning of the agent, however, UPIES is much faster. While LPIES/RPIES is slower than UPIES it does significantly better than the no-shaping agent. A key qualitative difference between the behavior of UPIES and LPIES/RPIES is that LPIES/RPIES pull the default RL and expert arms with equal probability whereas UPIES prefers the expert arm in this case. 

\subsection{Pong}
Next, we show that UPIES is able to accelerate the learning of an RL agent learning to play Pong directly from pixels. The advice in this case comes from another DQN that was trained to play Pong. We only try UPIES here to save computation and time. The agent gets a positive reward for imitating the expert agent. Figure \ref{fig:cartpole}(right) show the performance of UPIES against a no-shaping agent. It turns out that in this case, the behavior of UPIES was similar to RPIES mainly because even if the agent followed the expert advice, it was not necessarily able to end with winning the game eventually. Nonetheless, the UPIES agent is able to beat the default RL agent in terms of the number of episodes required to learn the optimal policy.
For Pong the expert and the RL agent both are a convolutional neural network with two convolutional layers of kernel size 8 and 4 respectively. The last layer is a dense layer of size 256. Rectified Linear Units (ReLU) are used as the activation on all three layers. The RL agent is trained with a learning rate of 0.00025 with a gamma of 0.99 with Adam optimizer. The expert reward was 0.01 for every time the RL agent took the action suggested by the expert agent

\section{Related Work}
The closest work to ours is that of PIES \citep{behboudian20}, two-level Q learning  \citep{li19twolevel}, and  incremental meta-gradient learning (IMLG) \citep{hu2020learning}. All three methods formulate the problem of shaping as a bi-level optimization problem. However, none of these methods address the challenge of non-stationarity resulting due to learning of underlying RL agents. As we show this can lead to sub-optimal results.
Before these methods, potential-based reward shaping \citep{harutyunyan15,devlin12} was commonly used for reward shaping. However, it is limited by its requirement of expressing the shaping advice as a potential function. 
Other related methods are learning to shape reward \citep{mguni2021learning} and heuristics-based RL \citep{cheng2021heuristic}. These methods do not directly address the challenge of incorporating shaping rewards but instead, address the challenge of how to express (or learn to express) a good shaping reward.
It is easy to see that unless any of these approaches take into account the non-stationarity of rewards they will fail in the two-armed bandit settings described earlier since they propose $\epsilon$-greedy policies that are greedy with respect to historically collected rewards.
PBRS is related to reward shaping and frameworks like TAMER \citep{knox2009interactively}. We show that PIES-based algorithms are able to learn from a teacher agent and accelerate learning. 
A big advantage of our method over most of these existing methods is that it is underlying RL model agnostic and simple to implement. One can take any underlying RL agent and simply apply our method on top of it. 

\section{Discussion and Future Work}
This paper presented a suite of methods for incorporating external advice in RL framework. The methods are adaptable to the needs of the user, simple to implement, and underlying RL model agnostic. We show that these methods in principle and in practice achieve the goals of reward shaping as outlined in \citep{behboudian20}. However, the propagation of uncertainty when computing the confidence bounds can be improved and weaker assumptions can help the results of this paper. Finally, another direction to extend the current method is to apply UPIES to every state-action pair at every time step instead of every episode as done in this paper. 

\section{Disclaimer}
This paper was prepared for informational purposes with contributions from the Global Technology Applied Research
center of JPMorgan Chase \& Co. This paper is not a product of the Research Department of JPMorgan Chase \& Co. or
its affiliates. Neither JPMorgan Chase \& Co. nor any of its affiliates makes any explicit or implied representation or
warranty and none of them accept any liability in connection with this paper, including, without limitation, with respect
to the completeness, accuracy, or reliability of the information contained herein and the potential legal, compliance,
tax, or accounting effects thereof. This document is not intended as investment research or investment advice, or as a
recommendation, offer, or solicitation for the purchase or sale of any security, financial instrument, financial product or
service, or to be used in any way for evaluating the merits of participating in any transaction.

\section{Acknowledgments}
The authors would like to thank Michael Bowling and Matthew Taylor for their constructive feedback. The authors would like to thank the anonymous reviewers at the adaptive and learning agents workshop for their constructive feedback. 

\bibliography{PB-bib}
\bibliographystyle{ACM-Reference-Format}

\end{document}